\documentclass[conference]{IEEEtran}
\IEEEoverridecommandlockouts
\usepackage{geometry}
\usepackage{amsmath,amssymb,amsfonts}
\usepackage{stfloats}
\usepackage{algorithmic}
\usepackage{graphicx}
\usepackage{textcomp}
\usepackage{xcolor}
\usepackage{multirow}
\usepackage{makecell}
\usepackage{booktabs}
\usepackage{subcaption}
\usepackage[backend=biber,sorting=none,style=ieee,url=false]{biblatex}
\usepackage{hyperref}
\usepackage{bm}
\usepackage{bbding}

\let\oldtwocolumn\twocolumn
\renewcommand\twocolumn[1][]{%
    \oldtwocolumn[{#1}{
    \begin{center}
           \includegraphics[width=\textwidth]{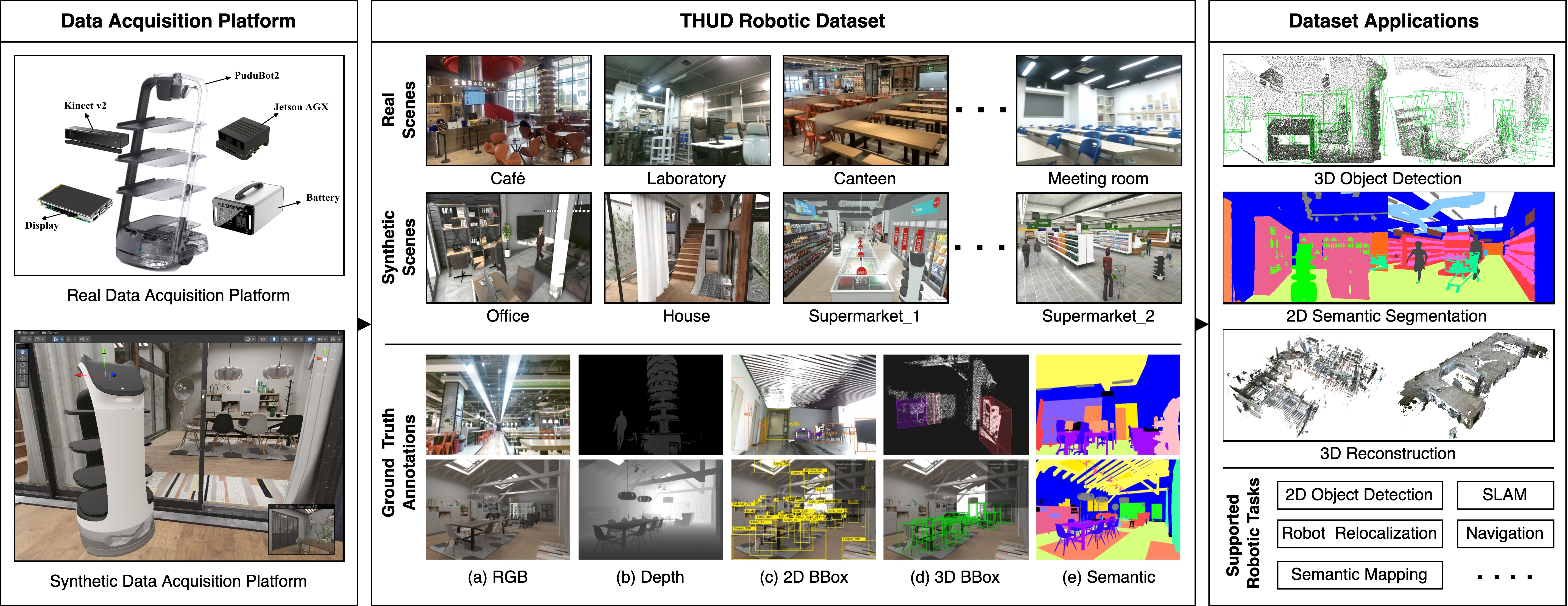}
           \captionof{figure}{THUD robotic dataset, first column: real and synthetic data acquisition platforms; second column: scene classes and annotation types; third column: applications supported by THUD.}
           \label{fig0}
        \end{center}
    }]
}

\begin{document}
\newgeometry{left=19.1mm,right=19.1mm,top=25.4mm,bottom=19.1mm}

\title{Mobile Robot Oriented Large-Scale Indoor Dataset for Dynamic Scene Understanding
\thanks{\textsuperscript{\dag}These authors contributed equally to this work.}
\thanks{\textsuperscript{*}Corresponding author.}
}

\author{Yi-Fan Tang$^{\dag}$, Cong Tai$^{\dag}$, Fang-Xing Chen$^{\dag}$, Wan-Ting Zhang, Tao Zhang, Xue-Ping Liu, \\ Yong-Jin Liu and Long Zeng$^{*}$
\\
Project website: \href{https://jackyzengl.github.io/THUD-Robotic-Dataset.github.io/}{https://jackyzengl.github.io/THUD-Robotic-Dataset.github.io/}
}
\maketitle

\begin{abstract}
Most existing robotic datasets capture static scene data and thus are limited in evaluating robots' dynamic performance. To address this, we present a mobile robot oriented large-scale indoor dataset, denoted as THUD (Tsinghua University Dynamic) robotic dataset, for training and evaluating their dynamic scene understanding algorithms. Specifically, the THUD dataset construction is first detailed, including organization, acquisition, and annotation methods. It comprises both real-world and synthetic data, collected with a real robot platform and a physical simulation platform, respectively. Our current dataset includes 13 larges-scale dynamic scenarios, 90K image frames, 20M 2D/3D bounding boxes of static and dynamic objects, camera poses, and IMU. The dataset is still continuously expanding. Then, the performance of mainstream indoor scene understanding tasks, e.g. 3D object detection, semantic segmentation, and robot relocalization, is evaluated on our THUD dataset. These experiments reveal serious challenges for some robot scene understanding tasks in dynamic scenes. By sharing this dataset, we aim to foster and iterate new mobile robot algorithms quickly for robot actual working dynamic environment, i.e. complex crowded dynamic scenes. 
\end{abstract}


\section{Introduction}
Mobile robots are widely used in various scenarios, e.g. restaurant and supermarket, which are typical dynamic environments with many moving objects. However, most existing mobile robotic datasets mainly capture data of static scenes, which cannot support well in training and evaluating mobile robots for their practical working status, particularly for large-scale indoor scenes. To overcome this limitation, we have constructed a dataset specifically designed for training mobile robot to perform various tasks in large-scale dynamic indoor scenes.

The overview of our mobile robot oriented large-scale indoor dataset, named THUD (Tsinghua University Dynamic), is shown in  Fig.~\ref{fig0}. First, the dataset is acquired in 13 dynamic scenarios, i.e. 8 real-world and 5 synthetic scenes, with moving pedestrians, mobile robots, shopping cart, and other dynamic objects. The data are collected via a real robot platform and a physical simulation platform, both containing different levels of dynamic complexity. This is useful in evaluating real mobile robots' performance in restaurant-like scenarios, which have vary dynamic complexity in different time period. Then, to provide rich information, the collected data is annotated with dense ground-truth labels,  including over 90K frames (each with a RGB image and a depth map), over 20M 2D/3D object detection bounding boxes, semantic segmentation annotations, and camera poses. This makes our dataset applicable to both static and dynamic mobile robot indoor tasks.
\restoregeometry
Thus, we evaluated and tested THUD on some representative algorithms from static and dynamic mobile robot indoor tasks, i.e. 3D object detection, semantic segmentation, and robot relocalization tasks. The results demonstrate that algorithms designed for different tasks exhibit varying degrees of performance degradation in scenes include dynamic objects, especially in robot relocalization. We can expect our dataset or its expanding version will support other static and dynamic indoor mobile robot tasks well, e.g. scene reconstruction, semantic map construction, robot navigation, object tracking, semantic segmentation, trajectory prediction, domain adaption, etc.

By utilizing the THUD dataset, we hope to provide a valuable resource for researchers and developers in the field of mobile robotics, fostering the development and performance improvement of mobile robot systems. By sharing this dataset, we aim to encourage more researchers to participate, facilitating continuous innovation and progress in mobile robot studies.

\section{Related Work}

\begin{table*}[t]
    \centering
    \caption{RGB-D Datasets Comparison}
    \label{RGB-D Datasets}
    \begin{tabular}{clcclccccccc}
    \toprule
    \multirow{2}*[-0.75ex]{\makecell{\textbf{Type}}} & \multirow{2}*[-0.75ex]{\textbf{Dataset}} & \multirow{2}*[-0.75ex]{\makecell{\textbf{Data}\\\textbf{type}}} & \multirow{2}*[-0.75ex]{\textbf{Year}} & \multirow{2}*[-0.75ex]{\textbf{\# Labels}} & \multirow{2}*[-0.75ex]{\makecell{\textbf{\# Annotations}\\\textbf{per frame}}} & \multirow{2}*[-0.75ex]{\makecell{\textbf{\# Object}\\\textbf{classes}}} & \multirow{2}*[-0.75ex]{\makecell{\textbf{Dynamic}\\\textbf{objects}$^a$}} & \multicolumn{4}{c}{\textbf{Tasks}$^b$} \\
    \cmidrule{9-12} 
    & & & & & & & & \textbf{\textit{2D}} & \textbf{\textit{3D}}& \textbf{\textit{SS}}& \textbf{\textit{RC}} \\
    \midrule
    \multirow{5}{*}{2D} & B3DO \cite{B3DO} & Real & 2011 & 849 frames & 2$\sim$5 & 50+ & × & \checkmark & × & × & × \\
    & NYU-Depth v2 \cite{NYU} & Real & 2012 & 1,449 frames & 30$\sim$40 & 894 & × & × & × & \checkmark & × \\
    & SUN3D \cite{SUN3D} & Real & 2013 & 8 scans & 10$\sim$15 & - & × & × & × & \checkmark & \checkmark \\
    & Stanford 2D-3D-S \cite{Stanford} & Real & 2017 & 70,496 frames & 10$\sim$15 & 13 & × & × & × & \checkmark & × \\
    & SceneNet RGB-D \cite{SceneNet} & Synthetic & 2017 & 5M frames & 20$\sim$30 & 255 & × & × & × & \checkmark & \checkmark \\
    \cmidrule{1-12}
    \multirow{8}{*}{3D} & SUN RGB-D \cite{SUNRGBD} & Real & 2015 & 10k frames & 20$\sim$30 & 800 & × & × & \checkmark & \checkmark & \checkmark \\
    & ScanNet \cite{ScanNet} & Real & 2017 & 2.5M frames & 10$\sim$15 & 21 & × & × & \checkmark & \checkmark & \checkmark \\
    & SUN-CG \cite{SUNCG} & Synthetic & 2017 & 500k frames & 5$\sim$15 & 84 & × & × & × & \checkmark & \checkmark \\
    & Matterport 3D \cite{Matterport3D} & Real & 2017 & 194,400 frames & 5$\sim$15 & 40 & × & × & × & \checkmark & \checkmark \\
    & InteriorNet \cite{InteriorNet} & Synthetic & 2018 & 20M frames & 20$\sim$30 & 158 & × & × & × & \checkmark & \checkmark \\
    & ARKitScenes \cite{ARKitScenes} & Real & 2022 & 5,047 scans & 5$\sim$10 & 17 & × & × & \checkmark & × & × \\
    & ScanNet++ \cite{ScanNet++} & Real & 2023 & 1,858 scans & 20$\sim$30 & 1,000+ & × & × & × & \checkmark & \checkmark \\
    \cmidrule{2-12}
    & \textbf{THUD(Ours)} & \textbf{Synthetic\&Real} & \textbf{2023} & \textbf{90k frames} & \textbf{150$\sim$200} & \textbf{91} & \checkmark & \checkmark & \checkmark & \checkmark & \checkmark \\
    \bottomrule
    \multicolumn{11}{l}{$^a$Dynamic objects: Includes moving robots, walking people, and people moving with shopping carts.} \\
    \multicolumn{11}{l}{$^b$2D: 2D Object Detection; 3D: 3D Object Detection; SS: Semantic Segmentation; RC: 3D Reconstruction.}
    \vspace{-10pt}
    \end{tabular}
\end{table*}

Scene understanding is one of the most fundamental topics in robot environment perception and includes many common tasks in computer vision, such as: 3D object detection, object classification, semantic segmentation, pose estimation, spatial layout estimation, CAD model retrieval and alignment, etc. With the development of more lightweight and low-cost RGB-D sensors, personal users can get convenient access to images and depth data of indoor scenes. Therefore, many related works on RGB-D datasets have emerged to meet the evolving needs of robot perception algorithms.

Some popular datasets are summarized in Table \ref{RGB-D Datasets}. Based on their annotation types, those RBG-D datasets are coarsely divided into two categories: annotated in 2D domain and annotated in 3D domain.

Since effective and intensive 3D annotations is hard to acquire, some works label RGB-D images with 2D annotations, which indirectly serve as ground truth for 3D tasks. Berkeley 3D Object Dataset \cite{B3DO} has 2D bounding box annotations on RGB-D images, NYU Depth v2 \cite{NYU} includes 2D semantic segmentation from short RGB-D videos with 1449 selected frames tagged, SUN3D \cite{SUN3D} dataset is composed of 415 RGB-D video sequences in 254 scenes. Stanford 2D-3D-Semantics dataset \cite{Stanford} utilizes the iGibson simulation environment to provide large scale virtual scenes with 2D texture, geometric as well as semantic information. SceneNet RGB-D dataset \cite{SceneNet} contains 5000k images and different kinds of 2D annotations.

3D annotation is challenging, yet some works have still made significant contributions. SUN RGB-D \cite{SUNRGBD} contains 10,335 RGB-D images with dense 2D/3D annotations, including 2D polygons, 3D bounding boxes with accurate object orientation and 3D room layout. ScanNet \cite{ScanNet} contains a total of 1,513 video sequences with annotations of 3D camera poses, surface reconstruction and semantic segmentation, and partially aligned CAD models. SUN-CG \cite{SUNCG} provides 45k virtual scene layouts and 500k ren- dered images with single-view RGB, depth maps and semantic segmentation maps. Matterport 3D \cite{Matterport3D} contains 194,400 RGB- D images to generate panoramas with surface reconstruction, camera position and 2D/3D semantic segmentation annota- tions. InteriorNet \cite{InteriorNet} is rendered entirely in virtual home scenes and contains 15k sequences. ARKitScenes \cite{ARKitScenes} improves the resolution of ground truth geometry from laser scans. ScanNet++ \cite{ScanNet++} is a new dataset that contains 460 high-resolution 3D reconstructions of indoor scenes with dense semantic and instance annotations.

Drawing references from both 2D and 3D annotation works, current datasets lack dynamic objects and pedestrians, which are highly common in real-world scenes \cite{dai_scannet_2017, armeni_joint_2017, hua_scenenn_2016, ICL-NUIM, song_suncgsemantic_2016, song_sun_2015, TUM, 12Scenes} and pose sufficient challenges for robot-related research. These datasets commonly exhibit issues such as poor data quality, small annotation volumes, limited label types, noisy annotations, and unreasonable virtual scene layouts. Current research often requires testing, training, and deployment on real hardware in virtual scenarios, which the current datasets cannot fully satisfy.

\subsection{Contributions and Advantages of THUD}
\begin{itemize}
\item Our dataset provides data annotated with dynamic instances for large-scale indoor scenes, which is closer to mobile robots' real working environment, posing significant challenges to dynamic robot tasks.
\item Our dataset supports training and testing for various robotic scene understanding tasks (object detection, semantic segmentation, robot relocalization, scene reconstruction, etc.)
\item Our dataset contains both real and synthetic annotated data, which can satisfy the testing of mobile robot algorithms in different scenes.
\end{itemize}

\section{Dataset Construction}
\subsection{Data Acquisition}

\subsubsection{Real Data Acquisition}
\textbf{}

\begin{figure}[h]
    \centering
    \includegraphics[width=8.5cm]{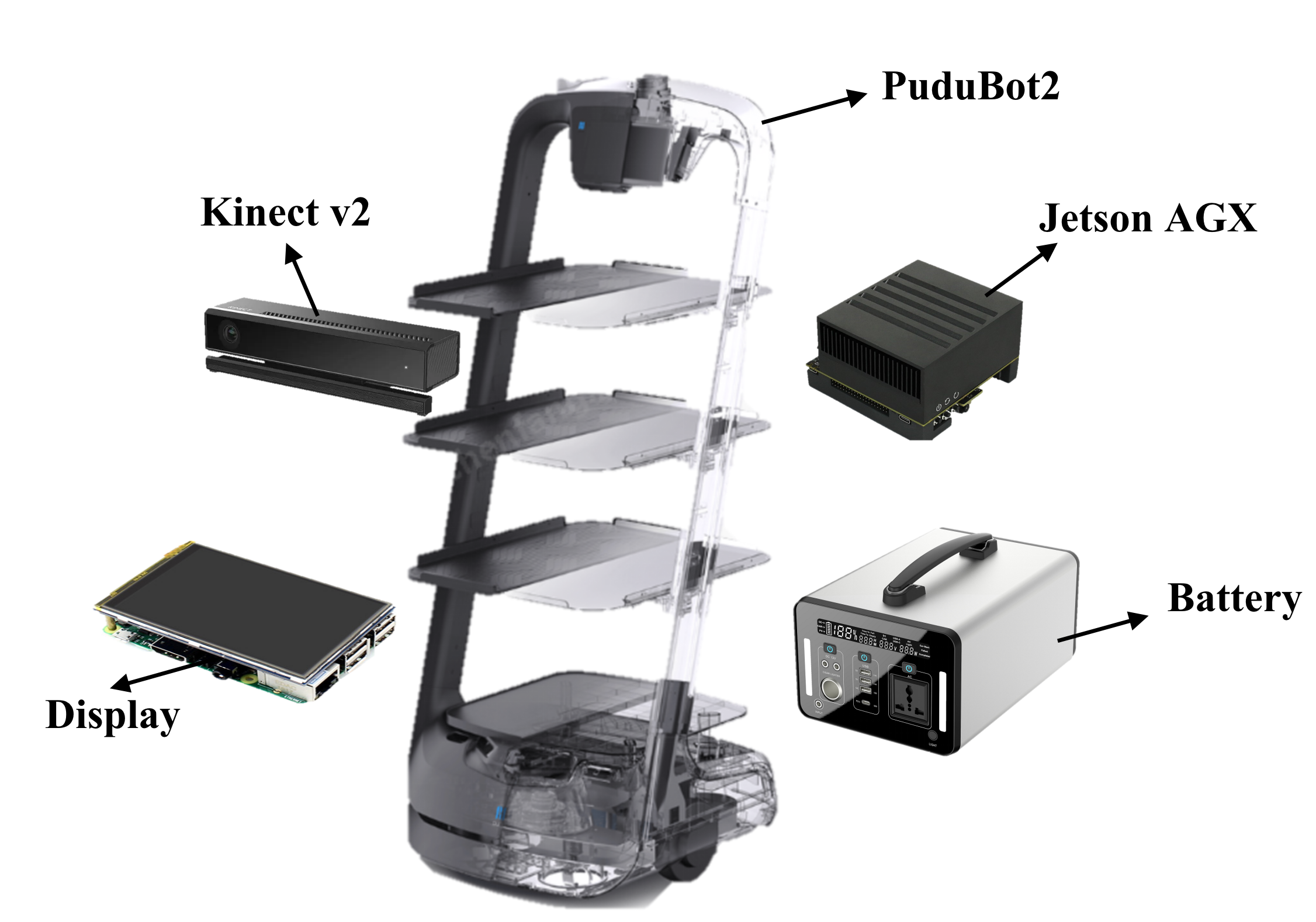}
    \caption{PUDUbot2 \& Kinect V2 joint collection platform}
    \vspace{-5pt}
    \label{fig2}
\end{figure}

\textbf{\textit{Acquisition platform:}}
Data collection relies on the PUDUbot2 \& Kinect V2 joint collection platform, as depicted in Fig.~\ref{fig2} PUDUbot2 is an advanced mobile service robot that, when in developer mode, can record real-time robot poses (translation distance and Euler angles). The depth and RGB images of real data are collected through the Kinect V2.
\vspace{-8pt}

\textbf{\textit{Acquisition scenarios:}}
The data collection for this dataset is primarily focused on the field of mobile robotics. In order to better serve this domain, the real-world data was collected in common service robot environments on the campus of Tsinghua University. As shown in Fig.\ref{fig0} colum 2nd. These include eight scenes: a corridor of laboratory, the lobby of teaching building, department meeting rooms, departmental visiting laboratories, a cafeteria, a campus dining hall, elevator area, and shops.

\begin{figure}[h]
    \centering
    \includegraphics[width=8.5cm]{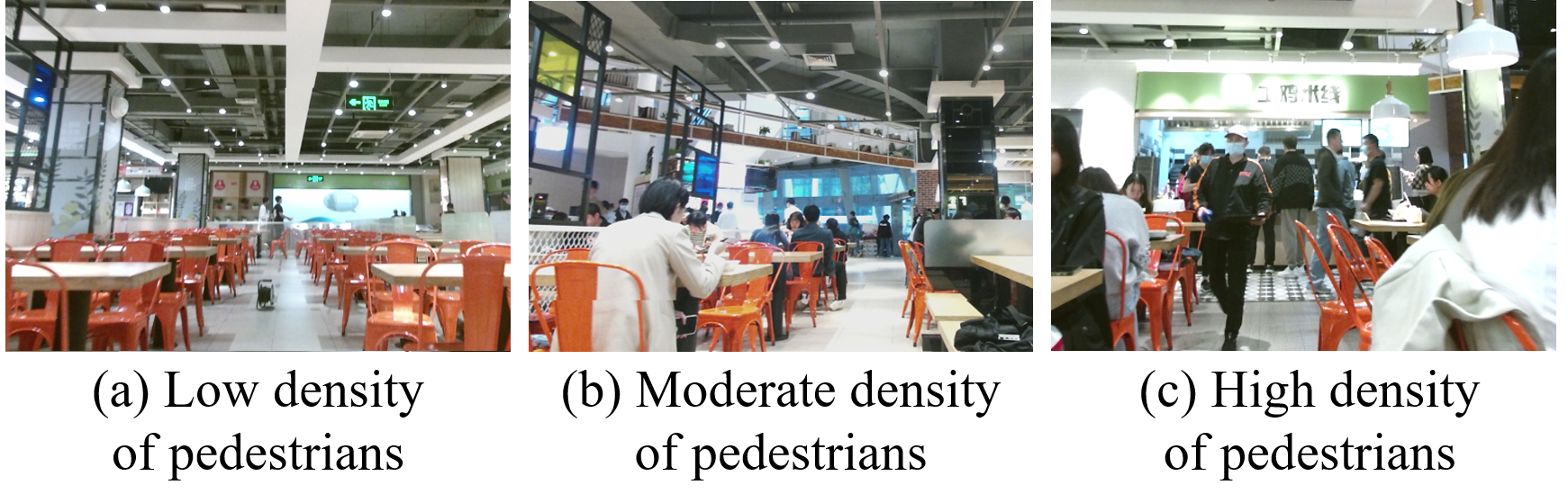}
    \caption{Scenes with varying levels of dynamic complexity}
    \vspace{-8pt}
    \label{fig3}
\end{figure}

One of the unique features of THUD is its dynamic nature. To showcase different levels of dynamic complexity in the real-world scenarios, data collection was conducted at various time intervals. For example, Fig.~\ref{fig3} illustrates the data collection process in the canteen scene at different time periods. Within different data sequences in the same scene, there are sequences with low, moderate and high density of pedestrians. These sequences with varying levels of dynamic complexity in the same environment will help users evaluate the algorithm's capabilities and better align with the real-world situations that service robots encounter at different times. Another notable feature of this dataset is the inclusion of various challenging labels, such as elevators, glass objects, and other item categories. These items have a significant impact on robot perception during actual testing.

\textbf{\textit{Acquisition method:}}
The real data collection system consists of two independent systems: pose acquisition and image acquisition. The acquisition of robot pose data primarily relies on the v-SLAM localization at the top of PUDUbot2, supplemented by the odometry of the steering wheel. The data is computed at a frame rate of 40fps.

The image data includes depth and RGB images. We have developed ROS scripts for synchronous acquisition of RGB and depth images. The depth images in the released version have been denoised using the Self-Supervised Deep Depth Denoising method \cite{Sterzentsenko_2019_ICCV}. Due to the inherent instability of Kinect V2, the actual frame rate for capturing RGB and depth images fluctuates between 15-30fps.

\subsubsection{Synthetic Data Acquisition}
\textbf{}

\begin{figure*}[ht]
    \centering
    \includegraphics[width=\linewidth]{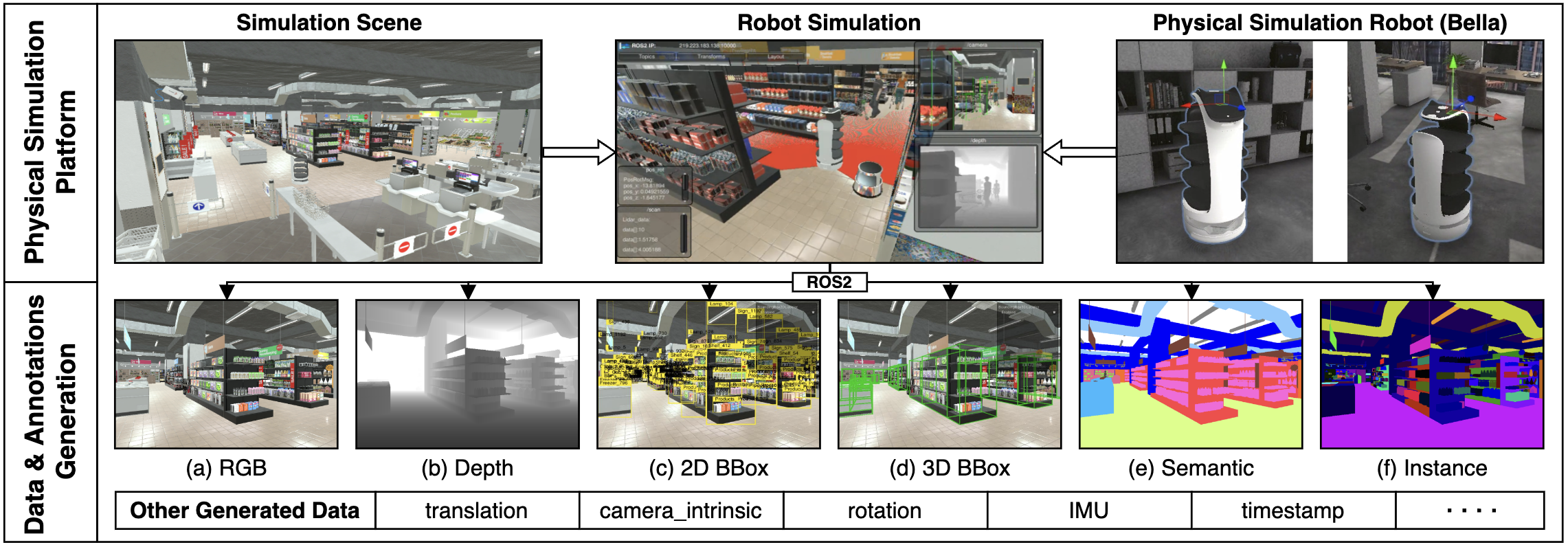}
    \caption{Mobile robot synthetic data acquisition platform.}
    \vspace{-10pt}
    \label{fig1}
\end{figure*}

\textbf{\textit{Acquisition platform:}}
The platform is designed to simulate the realistic working environment for mobile robots, allowing for the collection of virtual sensor data such as RGB images, depth maps, robot pose, and IMU data. The platform has been developed using open source tools from Unity3D and has undergone secondary development to meet the specific requirements of the synthetic data acquisition for mobile robots. The Unity3D physics engine is utilized to simulate the physical movements of the mobile robot, and simultaneously publish the collected synthetic data and labels through ROS2. The platform architecture is shown in Fig.~\ref{fig1}. The ultimate aim is to provide a comprehensive synthetic data acquisition platform that can be used to train and test mobile robot indoor scenes understanding algorithms.

\begin{figure}[h]
    \centering
    \includegraphics[width=\linewidth]{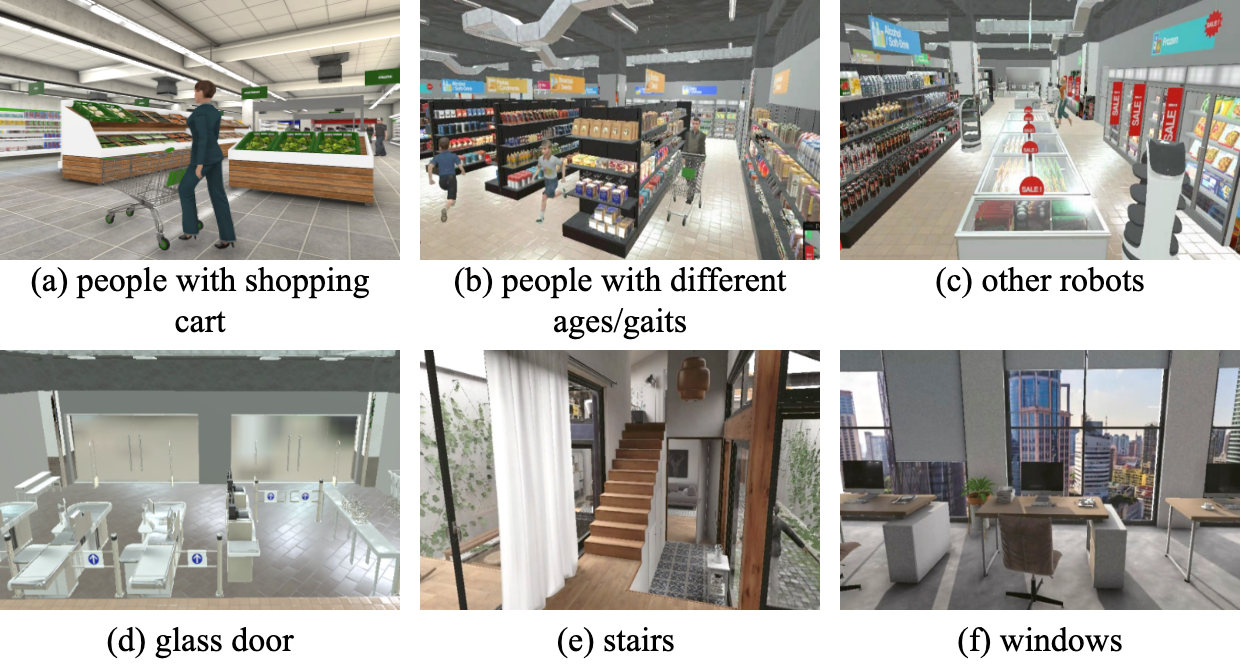}
    \caption{Dynamic and special scenarios}
    \vspace{-10pt}
    \label{fig:scenes}
\end{figure}

\textbf{\textit{Acquisition scenarios:}}
To acquire more realistic and effective data, the scenes were designed based on the actual working environment of mobile robots, taking into account two main aspects: dynamic scenarios with moving obstacles and special scenarios that pose potential hazards for robots in the reality. As illustrated in Fig.~\ref{fig:scenes}. Dynamic scenarios were created by setting up pedestrians in the scenes, some special cases were also considered, such as running children, people pushing shopping carts, other moving robots in the scene. The scenes also take into account pedestrians of different ages and clothing styles for men, women, and children. For special scenarios, the scene settings focus on realistic objects that tend to pose difficulties for robotic tasks, such as stairs, windows, glass doors, etc.

\textbf{\textit{Acquisition method:}}
The synthetic data is generated by creating virtual sensors in Unity3D scenes. Based on the type and position of sensors on the real mobile robot, the camera for the physical simulation robot is configured accordingly, with a height of 1.2m and a pitch angle of 0 degrees. As the robot moves through the scenes, \textit{RGB images, depth maps, robot poses,} and \textit{IMU} data are captured. The resolution of the RGB images and depth maps is \textit{730*530}, and the robot pose is represented by its \textit{xyz} coordinates in the scene and its rotation angle around the \textit{z-axis}.

\begin{figure*}[t]
    \centering
    \includegraphics[width=\linewidth]{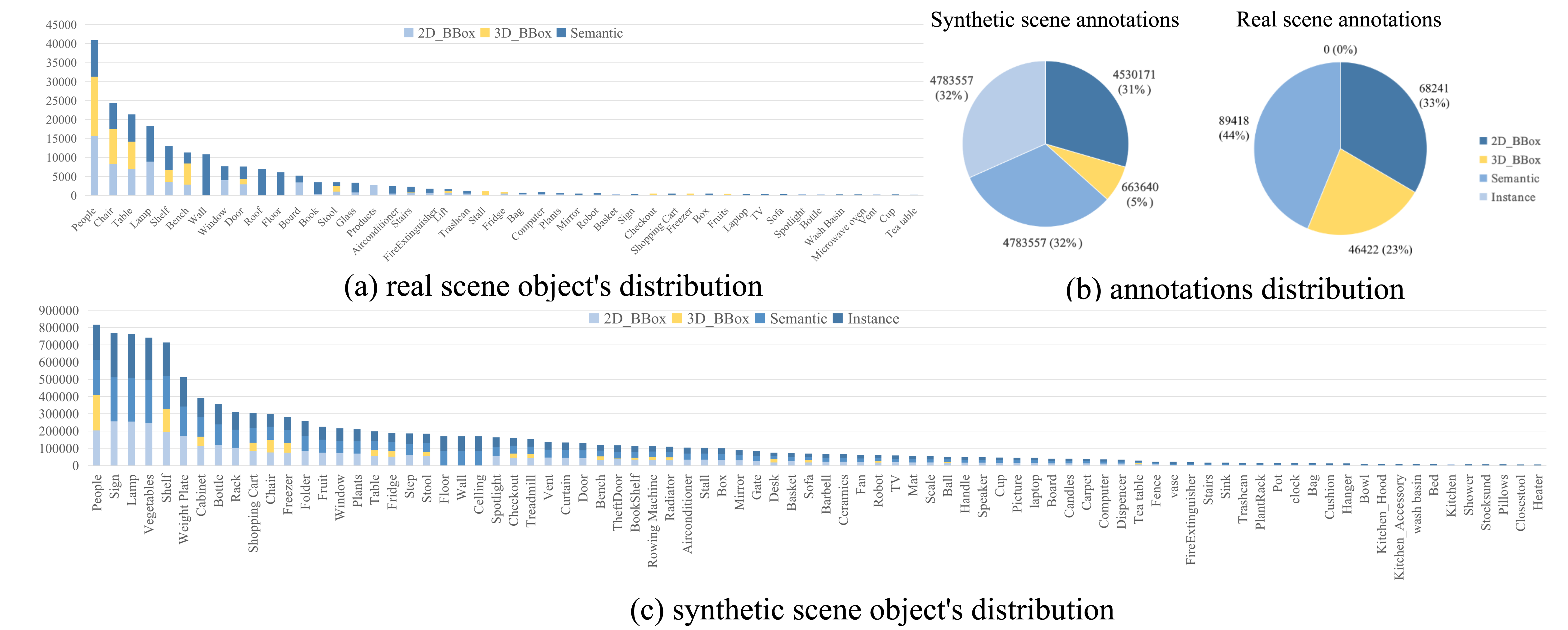}
    \caption{Statistics of annotations in our dataset.}
    \vspace{-8pt}
    \label{annotaion1}
\end{figure*}

\subsection{Ground Truth Annotation}
\subsubsection{Synthetic Data Annotation}
The annotation of the simulated data was automatically generated through our custom-built data acquisition platform. We integrated the robot model of \textit{Bella} into the virtual scenes for data collection, enabling the robot to navigate and capture data from its own perspective. During the data collection process, we simultaneously extracted vital information, including 2D and 3D bounding boxes, semantic and instance segmentation images, robot's pose and IMU information. These annotations were then synchronized with the captured RGB images and depth maps, ensuring a one-to-one correspondence.

Importantly, the annotation information extracted within our data acquisition platform represents ground truth within the virtual scenes. This ensures a high level of precision and accuracy in the generated annotations, allowing us to rapidly generate large quantities of meticulously annotated data within a short time.

\subsubsection{Real Data Annotation}
To annotate the real-world data with 2D and 3D object information, we first preprocessed RGB images and depth maps by applying automated algorithms to predict the 2D\&3D positions and bounding boxes of objects, such as Faster-RCNN \cite{FasterRCNN} and Votenet \cite{ImVoteNet}. However, the algorithms for 3D objects detection are not very effective. Subsequently, we performed manual review and verification to correct and refine the annotated bounding boxes, ensuring the accuracy and completeness of the detected objects. For the 2D semantic and instance segmentation task, we similarly employed a semi-automated annotation method, manual review and correction were also conducted by annotators to ensure accurate pixel-wise classification and semantic labeling.

In summary, our dataset construction process leveraged a combination of automated and semi-automated annotation methods to enhance efficiency and accuracy in data annotation.For providing high-quality training and evaluation data for robots indoor scene understanding tasks.

\subsection{Data Statistics}
Our dataset comprises a total of 90,175 annotated frames, consisting of 84,984 frames from synthetic data collection and 5,191 frames from real-world data collection. Each frame has undergone intensive annotation, resulting in a total of over 20M labels, with over 1.2M labels for dynamic objects such as pedestrians, robots, and shopping carts. On average, each frame contains 176 data labels.

These labels encompass four annotation types: 2D bounding boxes, 3D bounding boxes, semantic segmentation, and instance segmentation, with their respective proportions as shown in Fig.~\ref{annotaion1}(b). The dataset encompasses 8 real and 5 synthetic large-scale indoor scenes, with average area exceeding $300m^2$. There are 91 different object categories within these scenes, and their distribution, along with the counts of each annotation type, is illustrated in Fig.~\ref{annotaion1}(a)(c).

\section{Evaluation on THUD Robotic Dataset}

To verify whether our dynamic dataset is necessary and useful for mobile robots, three tasks, i.e. 3D object detection, semantic segmentation, and robot relocalization, are tested and analyzed. Important results and conclusion are given here and their analysis are elaborated in \textit{project website}.

\begin{table}[b]
\centering
\caption{Experiments on 3D Object Detection}
\label{tab:3D object detection results}
\begin{tabular}{cccc}
    \toprule
    \multirow{2}{*}{\textbf{Scene}}  & \multirow{2}{*}{\textbf{Method}} & \textbf{Dynamic Objs} & \textbf{Static Objs}   \\
     & & \textbf{(mAP)} & \textbf{(mAP)} \\
    \midrule
    \multirow{3}{*}{Supermarket} & F-PointNet \cite{F-PointNet} & 7.89 & 8.92\\& ImVoteNet \cite{ImVoteNet}  & 17.49 & 17.29 \\& DeMF \cite{DeMF}  & 34.51 & 38.24\\
    \midrule
    \multirow{3}{*}{Canteen} & F-PointNet \cite{F-PointNet} & 18.16 & 36.67\\& ImVoteNet \cite{ImVoteNet}  & 26.72 & 43.37 \\& DeMF \cite{DeMF}  & 28.56 & 45.43\\
    \bottomrule
    \vspace{-10pt}
\end{tabular}
\end{table}

\subsection{3D Object Detection}
3D object detection methods estimate a 3D bounding box and a pose of each object presented in a scene \cite{DeformableDETR, wang_tokenfusionmultimodal_2022, wang_frustum_2019}. To compare the performance difference, three representative indoor 3D object detection algorithms which have good performance on both dynamic and static objects, i.e. F-PointNet \cite{F-PointNet}, ImVoteNet \cite{ImVoteNet} and DeMF \cite{DeMF}, are selected. These algorithms were evaluated on the real-world canteen scene and the synthetic supermarket scene to assess their mean Average Precision (mAP) for static and dynamic objects. The experimental results are depicted in Table \ref{tab:3D object detection results}. In the supermarket scene, the selected 5 static objects include \textit{Chair, Table, Shelf, Cabinet, Fridge}. While the 3 dynamic objects encompass \textit{Shopping cart, Robot, People}. In the canteen scene, the 7 static objects are \textit{Chair, Table, Shelf, Bench, Door, Stairs, Stool}, and the dynamic object is only \textit{People}. Moreover, we used the average number of dynamic objects per frame as a metric to assess the level of scene dynamic complexity, the supermarket and canteen scene dynamic complexity is 0.94 and 3.34, respectively.

We can see two important conclusion from the test results in Table \ref{tab:3D object detection results}. First, the results indicate that different algorithms have varying degrees of decrease in accuracy when detecting dynamic objects as compared to static objects. Second, 
it can be observed that having a larger density of dynamic objects lead to more significant accuracy difference between static and dynamic objects. But this could also be influenced by the object categories.

\begin{table}[b]
\caption{Experiments on Semantic Segmentation}
\vspace{-8pt}
\begin{center}
\label{tab:SS_results}
\begin{tabular}{cccc}
    \toprule
    \textbf{Scene} & \textbf{Method} & \textbf{Backbone} & \textbf{MIoU(\%)}  \\
    \midrule
    \multirow{4}{*}{Supermarket} & ACNet \cite{ACNet} & 3×R50 & 74.83 \\
     & RedNet \cite{RedNet} & 2×R34 & 76.92 \\
     & ESANet \cite{ESANet} & 2×R34 & 78.42 \\
     & SA-Gate \cite{SA-Gate} & 2×R101 & 83.19 \\
    \midrule
    \multirow{4}{*}{Canteen} & ACNett \cite{ACNet} & 3×R50 & 51.85 \\
     & RedNet \cite{RedNet} & 2×R34 & 59.83 \\
     & ESANet \cite{ESANet} & 2×R34 & 65.97 \\
     & SA-Gate \cite{SA-Gate} & 2×R101 & 58.34 \\
    \bottomrule
    \vspace{-15pt}
\end{tabular}
\end{center}
\end{table}

\begin{figure*}[t]
    \centering
    \includegraphics[width=16cm, height=5.5cm]{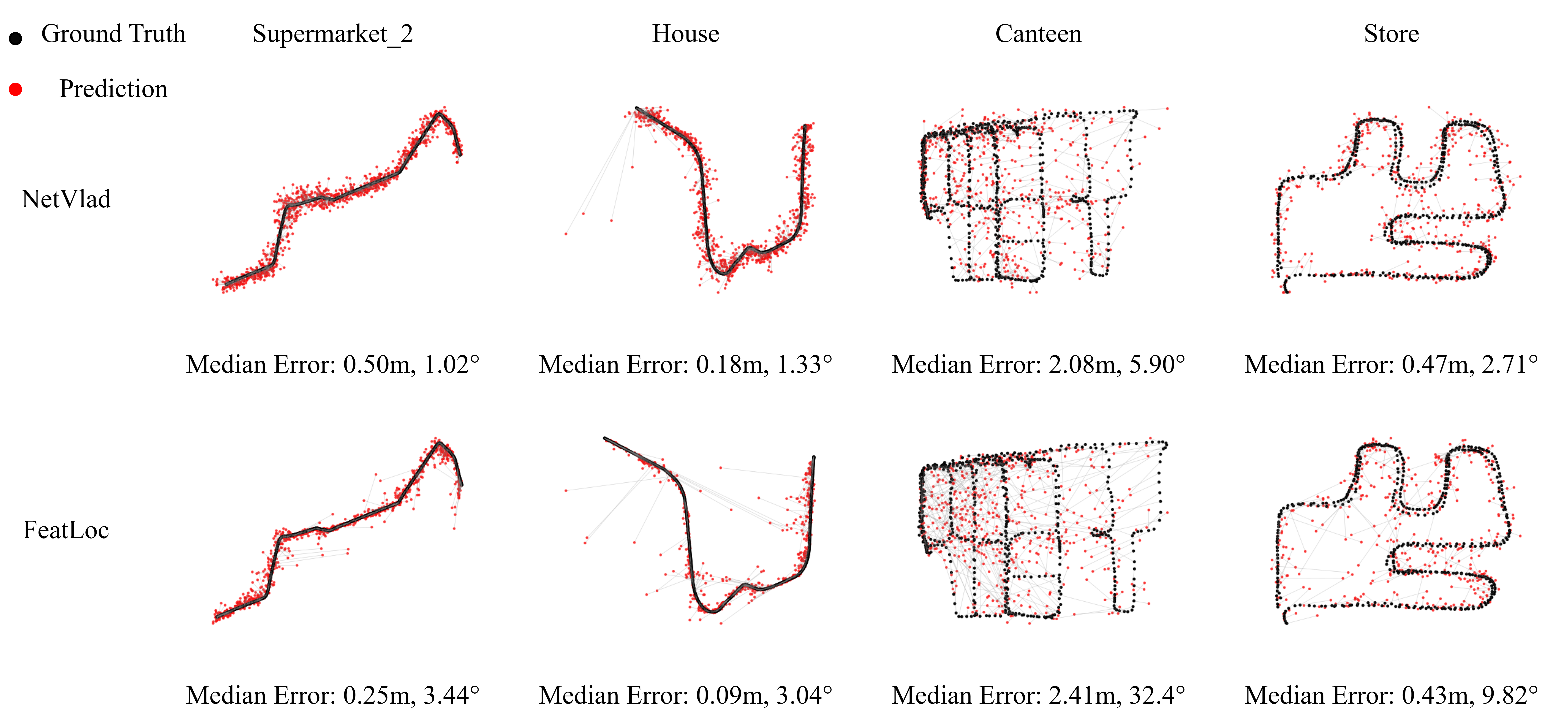}
    \caption{Test result visualization for robot relocalization}
    \label{fig_visualization}
    \vspace{-10pt}
\end{figure*}

\subsection{Semantic Segmentation}
Semantic segmentation methods assign each pixel in an image to its corresponding semantic category, focusing on the geometric structure and differentiation of different objects \cite{SS_survey, SS_survey_2, SS_survey_3}. However, robot indoor semantic segmentation faces several challenges, e.g. complex indoor scenes, occlusions, lighting variations, as well as the diversity of object shapes and sizes. Four RGB-D semantic segmentation algorithms, i.e. ACNet \cite{ACNet}, RedNet \cite{RedNet}, ESANet \cite{ESANet}, and SA-Gate \cite{SA-Gate}, are selected and tested on the aforementioned synthetic supermarket scene and real-world canteen scene. 31 and 19 static and dynamic objects with semantic labels are added to the supermarket and canteen scene, respectively. Objects without labels are set to the void value with corresponding pixels (0, 0, 0).

\begin{figure}[b]
    \centering
    \includegraphics[width=\linewidth]{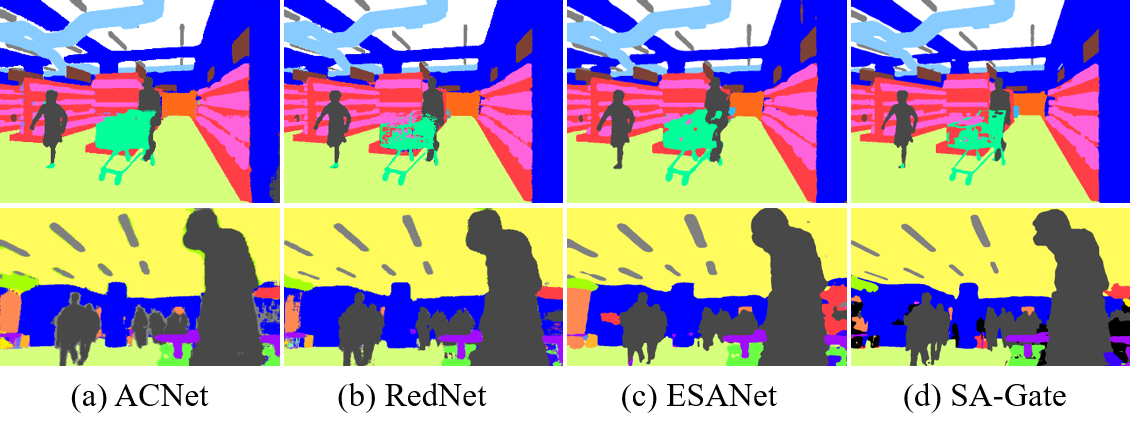}
    \caption{Comparison of semantic segmentation methods}
    \vspace{-10pt}
    \label{fig:SS_results}
\end{figure}

The accuracy of different semantic segmentation methods is measured by calculating the MIoU (Mean Intersection over Union) during testing and the results are shown in Table \ref{tab:SS_results}. We can obtain three conclusions. First, the accuracy on real-world canteen scene is much lower than synthetic supermarket scene. Second, the MIoU results are comparable to their original published results, though tested on different datasets. Third, to obtain a quantitative comparison on dynamic objects, we also train and test the ESANet \cite{ESANet} algorithm in the supermarket scene with and without dynamic objects, the MIoU of the test is 78.42\% and 79.63\% respectively. There is no significant difference, which is coincident with our intuition. In addition, as shown in Fig.~\ref{fig:SS_results}, different methods still have varying segmentation ability on dynamic object, such as the people and shopping carts.

\subsection{Robot Relocalization}
Robot relocalization refers to the process of accurately determines and adjusts its position based on the environmental perception information and its own localization capabilities, during its movement \cite{posenet, torii2013visual,dfnet}. Existing methods can be coarsely classified into global- and local-feature based methods.

We conducted extensive experiments on THUD using a representative global- and local-feature based methods, i.e. NetVlad \cite{NetVLAD} and FeatLoc \cite{FeatLoc}, respectively. We randomly selected two sequences as testing and the results on both synthetic and real-world scene are visualized in Fig.~\ref{fig_visualization}. 

\begin{figure}[h]
    \centering
    \includegraphics[width=\linewidth]{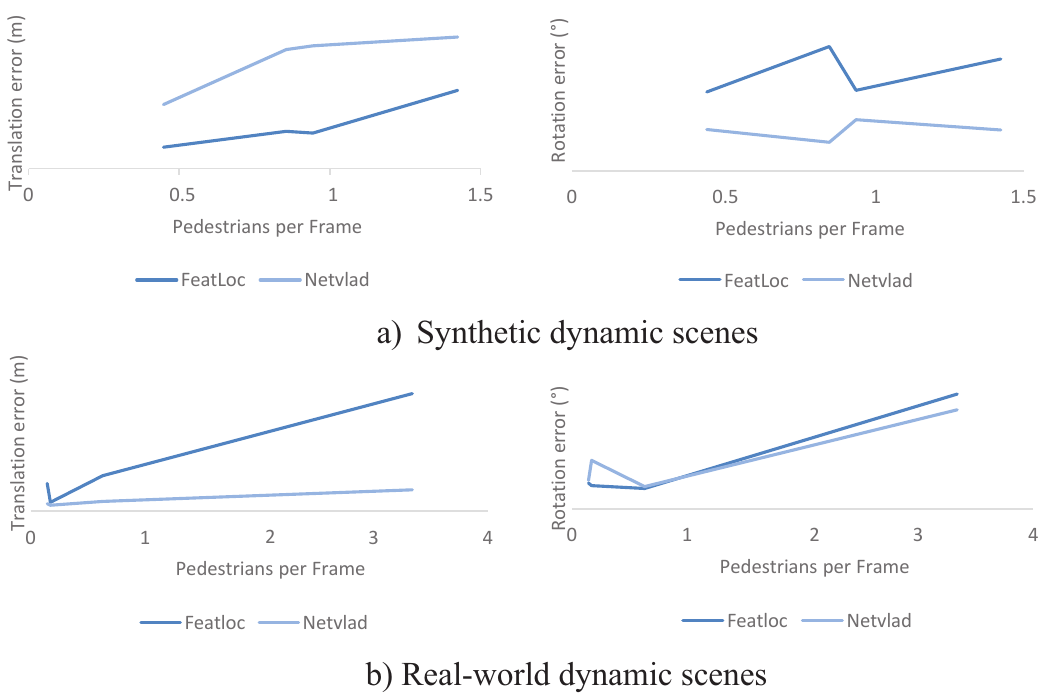}
    \caption{Trans.\& rot. error with different dynamic complexity}
    \label{fig_relocalization_dynamic}
\end{figure}

In addition, to analyze the effect of the number of dynamic objects (primarily pedestrians) in various scene images, we define a metric to represent a scene's dynamic complexity, computed as the average number of dynamic pedestrians per frame across the entire dataset. As shown in Fig.~\ref{fig_relocalization_dynamic}, with the scene's dynamic complexity increasing, the methods' accuracy decreases and  FeatLoc \cite{FeatLoc} has a higher drop in accuracy compared to NetVlad \cite{NetVLAD}, attributed to dynamic objects introducing disturbances and noise in the extraction of scene image features.

\section{Conclusions}
In this paper, we introduced a mobile robot oriented large-scale indoor dataset for dynamic scene understand tasks. It consists of both real and synthetic data to support both static and dynamic robotic task. In the near future, we are continuously expanding and iterating this dataset, to cover more robot dynamic tasks, e.g. robot navigation, object tracking, trajectory prediction, etc. The dataset initial version is shared at: \href{https://jackyzengl.github.io/THUD-Robotic-Dataset.github.io/}{https://jackyzengl.github.io/THUD-Robotic-Dataset.github.io/.}

\newpage
\printbibliography

@inproceedings{dfnet,
  title={Dfnet: Enhance absolute pose regression with direct feature matching},
  author={Chen, Shuai and Li, Xinghui and Wang, Zirui and Prisacariu, Victor A},
  booktitle={European Conference on Computer Vision},
  pages={1--17},
  year={2022},
  organization={Springer}
}

@inproceedings{torii2013visual,
  title={Visual place recognition with repetitive structures},
  author={Torii, Akihiko and Sivic, Josef and Pajdla, Tomas and Okutomi, Masatoshi},
  booktitle={Proceedings of the IEEE conference on computer vision and pattern recognition},
  pages={883--890},
  year={2013}
}

@inproceedings{posenet,
  title={Posenet: A convolutional network for real-time 6-dof camera relocalization},
  author={Kendall, Alex and Grimes, Matthew and Cipolla, Roberto},
  booktitle={Proceedings of the IEEE international conference on computer vision},
  pages={2938--2946},
  year={2015}
}

@InProceedings{Sterzentsenko_2019_ICCV,
author = {Sterzentsenko, Vladimiros and Saroglou, Leonidas and Chatzitofis, Anargyros and Thermos, Spyridon and Zioulis, Nikolaos and Doumanoglou, Alexandros and Zarpalas, Dimitrios and Daras, Petros},
title = {Self-Supervised Deep Depth Denoising},
booktitle = {Proceedings of the IEEE/CVF International Conference on Computer Vision (ICCV)},
month = {October},
year = {2019}
}

@INPROCEEDINGS{SUN3D,
  author={Xiao, Jianxiong and Owens, Andrew and Torralba, Antonio},
  booktitle={2013 IEEE International Conference on Computer Vision}, 
  title={SUN3D: A Database of Big Spaces Reconstructed Using SfM and Object Labels}, 
  year={2013},
  pages={1625-1632}}

@inproceedings{NYU,
  author    = {Nathan Silberman, Derek Hoiem, Pushmeet Kohli and Rob Fergus},
  title     = {Indoor Segmentation and Support Inference from RGBD Images},
  booktitle = {ECCV},
  year      = {2012}
}

@INPROCEEDINGS{B3DO,
  author={Janoch, Allison and Karayev, Sergey and Yangqing Jia and Barron, Jonathan T. and Fritz, Mario and Saenko, Kate and Darrell, Trevor},
  booktitle={2011 IEEE International Conference on Computer Vision Workshops (ICCV Workshops)}, 
  title={A category-level 3-D object dataset: Putting the Kinect to work}, 
  year={2011},
  volume={},
  number={},
  pages={1168-1174},
  doi={10.1109/ICCVW.2011.6130382}}

@article{Stanford,
  title={Joint 2D-3D-Semantic Data for Indoor Scene Understanding},
  author={Iro Armeni and Sasha Sax and Amir Roshan Zamir and Silvio Savarese},
  journal={ArXiv},
  year={2017},
  volume={abs/1702.01105}
}

@article{SceneNet,
  title={SceneNet RGB-D: 5M Photorealistic Images of Synthetic Indoor Trajectories with Ground Truth},
  author={John McCormac and Ankur Handa and Stefan Leutenegger and Andrew J. Davison},
  journal={ArXiv},
  year={2016},
  volume={abs/1612.05079}
}

@INPROCEEDINGS{SUNRGBD,
  author={Song, Shuran and Lichtenberg, Samuel P. and Xiao, Jianxiong},
  booktitle={2015 IEEE Conference on Computer Vision and Pattern Recognition (CVPR)}, 
  title={SUN RGB-D: A RGB-D scene understanding benchmark suite}, 
  year={2015},
  volume={},
  number={},
  pages={567-576},
  doi={10.1109/CVPR.2015.7298655}}

@INPROCEEDINGS{ScanNet,
  author={Dai, Angela and Chang, Angel X. and Savva, Manolis and Halber, Maciej and Funkhouser, Thomas and Nießner, Matthias},
  booktitle={2017 IEEE Conference on Computer Vision and Pattern Recognition (CVPR)}, 
  title={ScanNet: Richly-Annotated 3D Reconstructions of Indoor Scenes}, 
  year={2017},
  volume={},
  number={},
  pages={2432-2443},
  doi={10.1109/CVPR.2017.261}}

@article{InteriorNet,
  title={InteriorNet: Mega-scale Multi-sensor Photo-realistic Indoor Scenes Dataset},
  author={Wenbin Li and Sajad Saeedi and John McCormac and Ronald Clark and Dimos Tzoumanikas and Qing Ye and Yuzhong Huang and R. Tang and Stefan Leutenegger},
  journal={ArXiv},
  year={2018},
  volume={abs/1809.00716}}

@INPROCEEDINGS{SUNCG,
  author={Song, Shuran and Yu, Fisher and Zeng, Andy and Chang, Angel X. and Savva, Manolis and Funkhouser, Thomas},
  booktitle={2017 IEEE Conference on Computer Vision and Pattern Recognition (CVPR)}, 
  title={Semantic Scene Completion from a Single Depth Image}, 
  year={2017},
  volume={},
  number={},
  pages={190-198},
  doi={10.1109/CVPR.2017.28}}

@INPROCEEDINGS{TUM,
  author={Sturm, Jürgen and Engelhard, Nikolas and Endres, Felix and Burgard, Wolfram and Cremers, Daniel},
  booktitle={2012 IEEE/RSJ International Conference on Intelligent Robots and Systems}, 
  title={A benchmark for the evaluation of RGB-D SLAM systems}, 
  year={2012},
  volume={},
  number={},
  pages={573-580},
  doi={10.1109/IROS.2012.6385773}}

@article{Matterport3D,
  title={Matterport3D: Learning from RGB-D Data in Indoor Environments},
  author={Angel X. Chang and Angela Dai and Thomas A. Funkhouser and Maciej Halber and Matthias Nie{\ss}ner and Manolis Savva and Shuran Song and Andy Zeng and Yinda Zhang},
  journal={2017 International Conference on 3D Vision (3DV)},
  year={2017},
  pages={667-676}
}

@article{ScanNet++,
  title={ScanNet++: A High-Fidelity Dataset of 3D Indoor Scenes},
  author={Chandan Yeshwanth and Yueh-Cheng Liu and Matthias Nie{\ss}ner and Angela Dai},
  journal={ArXiv},
  year={2023},
  volume={abs/2308.11417}}

@inproceedings{ARKitScenes,
  title={ARKitScenes: A Diverse Real-World Dataset For 3D Indoor Scene Understanding Using Mobile RGB-D Data},
  author={Afshin Dehghan and Gilad Baruch and Zhuoyuan Chen and Yuri Feigin and Peter Fu and Thomas Gebauer and Daniel Kurz and Tal Dimry and Brandon Joffe and Arik Schwartz and Elad Shulman},
  booktitle={NeurIPS Datasets and Benchmarks},
  year={2021}}

@INPROCEEDINGS{ICL-NUIM,
  author={Handa, Ankur and Whelan, Thomas and McDonald, John and Davison, Andrew J.},
  booktitle={2014 IEEE International Conference on Robotics and Automation (ICRA)}, 
  title={A benchmark for RGB-D visual odometry, 3D reconstruction and SLAM}, 
  year={2014},
  volume={},
  number={},
  pages={1524-1531},
  doi={10.1109/ICRA.2014.6907054}}

@INPROCEEDINGS{F-PointNet,
  author={Qi, Charles R. and Liu, Wei and Wu, Chenxia and Su, Hao and Guibas, Leonidas J.},
  booktitle={2018 IEEE/CVF Conference on Computer Vision and Pattern Recognition}, 
  title={Frustum PointNets for 3D Object Detection from RGB-D Data}, 
  year={2018},
  volume={},
  number={},
  pages={918-927},
  doi={10.1109/CVPR.2018.00102}}

@INPROCEEDINGS{ImVoteNet,
  author={Qi, Charles R. and Chen, Xinlei and Litany, Or and Guibas, Leonidas J.},
  booktitle={2020 IEEE/CVF Conference on Computer Vision and Pattern Recognition (CVPR)}, 
  title={ImVoteNet: Boosting 3D Object Detection in Point Clouds With Image Votes}, 
  year={2020},
  volume={},
  number={},
  pages={4403-4412},
  doi={10.1109/CVPR42600.2020.00446}}

@ARTICLE{FasterRCNN,
  author={Ren, Shaoqing and He, Kaiming and Girshick, Ross and Sun, Jian},
  journal={IEEE Transactions on Pattern Analysis and Machine Intelligence}, 
  title={Faster R-CNN: Towards Real-Time Object Detection with Region Proposal Networks}, 
  year={2017},
  volume={39},
  number={6},
  pages={1137-1149},
  doi={10.1109/TPAMI.2016.2577031}}

@article{DeMF,
  title={Boosting 3D Object Detection via Object-Focused Image Fusion},
  author={Hao Yang and Chen Shi and Yihong Chen and Liwei Wang},
  journal={ArXiv},
  year={2022},
  volume={abs/2207.10589}}

@article{DeformableDETR,
  title={Deformable DETR: Deformable Transformers for End-to-End Object Detection},
  author={Xizhou Zhu and Weijie Su and Lewei Lu and Bin Li and Xiaogang Wang and Jifeng Dai},
  journal={ArXiv},
  year={2020},
  volume={abs/2010.04159}}

@INPROCEEDINGS{SS_survey,
  author={Cao, Fude and Bao, Qinghai},
  booktitle={2020 International Conference on Communications, Information System and Computer Engineering (CISCE)}, 
  title={A Survey On Image Semantic Segmentation Methods With Convolutional Neural Network}, 
  year={2020},
  volume={},
  number={},
  pages={458-462}}

@INPROCEEDINGS{SS_survey_2,
  author={Hu, Yaosi and Chen, Zhenzhong and Lin, Weiyao},
  booktitle={2018 IEEE International Conference on Multimedia \& Expo Workshops (ICMEW)}, 
  title={RGB-D Semantic Segmentation: A Review}, 
  year={2018},
  volume={},
  number={},
  pages={1-6}}

@article{SS_survey_3,
  title={A brief survey on RGB-D semantic segmentation using deep learning},
  author={Changshuo Wang and Chen Wang and Weijun Li and Haining Wang},
  journal={Displays},
  year={2021},
  volume={70},
  pages={102080}}

@article{RedNet,
  title={RedNet: Residual Encoder-Decoder Network for indoor RGB-D Semantic Segmentation},
  author={Jindong Jiang and Lunan Zheng and Fei Luo and Zhijun Zhang},
  journal={ArXiv},
  year={2018},
  volume={abs/1806.01054}
}

@inproceedings{SA-Gate,
  title={Bi-directional Cross-Modality Feature Propagation with Separation-and-Aggregation Gate for RGB-D Semantic Segmentation},
  author={Xiaokang Chen and Kwan-Yee Lin and Jingbo Wang and Wayne Wu and Chen Qian and Hongsheng Li and Gang Zeng},
  booktitle={European Conference on Computer Vision},
  year={2020}
}

@INPROCEEDINGS{ACNet,
  author={Hu, Xinxin and Yang, Kailun and Fei, Lei and Wang, Kaiwei},
  booktitle={2019 IEEE International Conference on Image Processing (ICIP)}, 
  title={ACNET: Attention Based Network to Exploit Complementary Features for RGBD Semantic Segmentation}, 
  year={2019},
  volume={},
  number={},
  pages={1440-1444}}

@article{ESANet,
  title={Efficient RGB-D Semantic Segmentation for Indoor Scene Analysis},
  author={Daniel Seichter and Mona K{\"o}hler and Benjamin Lewandowski and Tim Wengefeld and Horst-Michael Gro{\ss}},
  journal={2021 IEEE International Conference on Robotics and Automation (ICRA)},
  year={2020},
  pages={13525-13531}}

@ARTICLE{NetVLAD,
  author={Arandjelović, Relja and Gronat, Petr and Torii, Akihiko and Pajdla, Tomas and Sivic, Josef},
  journal={IEEE Transactions on Pattern Analysis and Machine Intelligence}, 
  title={NetVLAD: CNN Architecture for Weakly Supervised Place Recognition}, 
  year={2018},
  volume={40},
  number={6},
  pages={1437-1451},
  doi={10.1109/TPAMI.2017.2711011}}

@article{FeatLoc,
  title={FeatLoc: Absolute pose regressor for indoor 2D sparse features with simplistic view synthesizing},
  author={Thuan Bui Bach and Tuan Tran Dinh and Joo Ho Lee},
  journal={ISPRS Journal of Photogrammetry and Remote Sensing},
  year={2022}}

@INPROCEEDINGS{12Scenes,
  author={Valentin, Julien and Dai, Angela and Niessner, Matthias and Kohli, Pushmeet and Torr, Philip and Izadi, Shahram and Keskin, Cem},
  booktitle={2016 Fourth International Conference on 3D Vision (3DV)}, 
  title={Learning to Navigate the Energy Landscape}, 
  year={2016},
  volume={},
  number={},
  pages={323-332},
  doi={10.1109/3DV.2016.41}}

@misc{wang_tokenfusionmultimodal_2022,
	title = {{TokenFusion}:Multimodal Token Fusion for Vision Transformers},
	url = {http://arxiv.org/abs/2204.08721},
	number = {{arXiv}:2204.08721},
	publisher = {{arXiv}},
	author = {Wang, Yikai and Chen, Xinghao and Cao, Lele and Huang, Wenbing and Sun, Fuchun and Wang, Yunhe},
	urldate = {2022-09-10},
	date = {2022-07-15},
	langid = {english},
	eprinttype = {arxiv},
	eprint = {2204.08721 [cs]},
}

@misc{wang_frustum_2019,
	title = {Frustum {ConvNet}: Sliding Frustums to Aggregate Local Point-Wise Features for Amodal 3D Object Detection},
	url = {http://arxiv.org/abs/1903.01864},
	shorttitle = {Frustum {ConvNet}},
	number = {{arXiv}:1903.01864},
	publisher = {{arXiv}},
	author = {Wang, Zhixin and Jia, Kui},
	urldate = {2022-10-30},
	date = {2019-08-14},
	langid = {english},
	eprinttype = {arxiv},
	eprint = {1903.01864 [cs]},
}

@inproceedings{song_sun_2015,
	address = {Boston, MA, USA},
	title = {{SUN} {RGB}-{D}: {A} {RGB}-{D} scene understanding benchmark suite},
	isbn = {978-1-4673-6964-0},
	shorttitle = {{SUN} {RGB}-{D}},
	url = {http://ieeexplore.ieee.org/document/7298655/},
	doi = {10.1109/CVPR.2015.7298655},
	language = {en},
	urldate = {2022-10-21},
	booktitle = {2015 {IEEE} {Conference} on {Computer} {Vision} and {Pattern} {Recognition} ({CVPR})},
	publisher = {IEEE},
	author = {Song, Shuran and Lichtenberg, Samuel P. and Xiao, Jianxiong},
	month = jun,
	year = {2015},
	pages = {567--576},
}

@misc{dai_scannet_2017,
	title = {{ScanNet}: {Richly}-annotated {3D} {Reconstructions} of {Indoor} {Scenes}},
	shorttitle = {{ScanNet}},
	url = {http://arxiv.org/abs/1702.04405},
	language = {en},
	urldate = {2022-11-02},
	publisher = {arXiv},
	author = {Dai, Angela and Chang, Angel X. and Savva, Manolis and Halber, Maciej and Funkhouser, Thomas and Nie{\ss}ner, Matthias},
	month = apr,
	year = {2017},
	note = {arXiv:1702.04405 [cs]},
}

@misc{song_suncgsemantic_2016,
	title = {{SUNCG}?{Semantic} {Scene} {Completion} from a {Single} {Depth} {Image}},
	url = {http://arxiv.org/abs/1611.08974},
	language = {en},
	urldate = {2022-11-02},
	publisher = {arXiv},
	author = {Song, Shuran and Yu, Fisher and Zeng, Andy and Chang, Angel X. and Savva, Manolis and Funkhouser, Thomas},
	month = nov,
	year = {2016},
	note = {arXiv:1611.08974 [cs]},
}

@misc{armeni_joint_2017,
	title = {Joint {2D}-{3D}-{Semantic} {Data} for {Indoor} {Scene} {Understanding}},
	url = {http://arxiv.org/abs/1702.01105},
	language = {en},
	urldate = {2022-11-10},
	publisher = {arXiv},
	author = {Armeni, Iro and Sax, Sasha and Zamir, Amir R. and Savarese, Silvio},
	month = apr,
	year = {2017},
	note = {arXiv:1702.01105 [cs]},
}

@inproceedings{hua_scenenn_2016,
	address = {Stanford, CA, USA},
	title = {{SceneNN}: {A} {Scene} {Meshes} {Dataset} with {aNNotations}},
	isbn = {978-1-5090-5407-7},
	shorttitle = {{SceneNN}},
	url = {http://ieeexplore.ieee.org/document/7785081/},
	doi = {10.1109/3DV.2016.18},
	language = {en},
	urldate = {2022-11-10},
	booktitle = {2016 {Fourth} {International} {Conference} on {3D} {Vision} ({3DV})},
	publisher = {IEEE},
	author = {Hua, Binh-Son and Pham, Quang-Hieu and Nguyen, Duc Thanh and Tran, Minh-Khoi and Yu, Lap-Fai and Yeung, Sai-Kit},
	month = oct,
	year = {2016},
	pages = {92--101},
}

\end{document}